%% file: main.tex
\documentclass[10pt,twocolumn,letterpaper]{article}

\usepackage{cvpr}              %

\input{preamble}
\input{math_commands.tex}

\usepackage{multirow}
\usepackage{pifont}   %
\usepackage{wrapfig}  %
\usepackage{adjustbox}
\usepackage{float}
\usepackage{amsmath}

\usepackage[most]{tcolorbox}
\usepackage{minted}
\definecolor{grenn_1}{RGB}{143, 255, 221} %
\definecolor{grenn_2}{RGB}{88, 201, 185} %

\newcommand{\cmark}{\ding{51}}
\newcommand{\xmark}{\ding{55}}

\definecolor{cvprblue}{rgb}{0.21,0.49,0.74}
\usepackage[pagebackref,breaklinks,colorlinks,allcolors=cvprblue]{hyperref}

\title{\methodName{}: Part-level Human Motion Generation and Composition}

\author{
\begin{tabular}{ccccc}
Chuqiao Li\textsuperscript{1} & \quad Xianghui Xie\textsuperscript{1,2} & \quad Yong Cao\textsuperscript{1} & \quad Andreas Geiger\textsuperscript{1} & \quad Gerard Pons-Moll\textsuperscript{1,2}
\end{tabular}
\vspace{-4mm}
\\ \\ 
{\small \textsuperscript{1}Tübingen AI Center, University of Tübingen, Germany}\\
{\small \textsuperscript{2}Max Planck Institute for Informatics, Saarland Informatics Campus, Germany}\\
{\small \url{https://coral79.github.io/frankenmotion/} }
}

\input{definitions}
\begin{document}

\twocolumn[{
\maketitle
\begin{center}
    \includegraphics[width=0.98\textwidth]{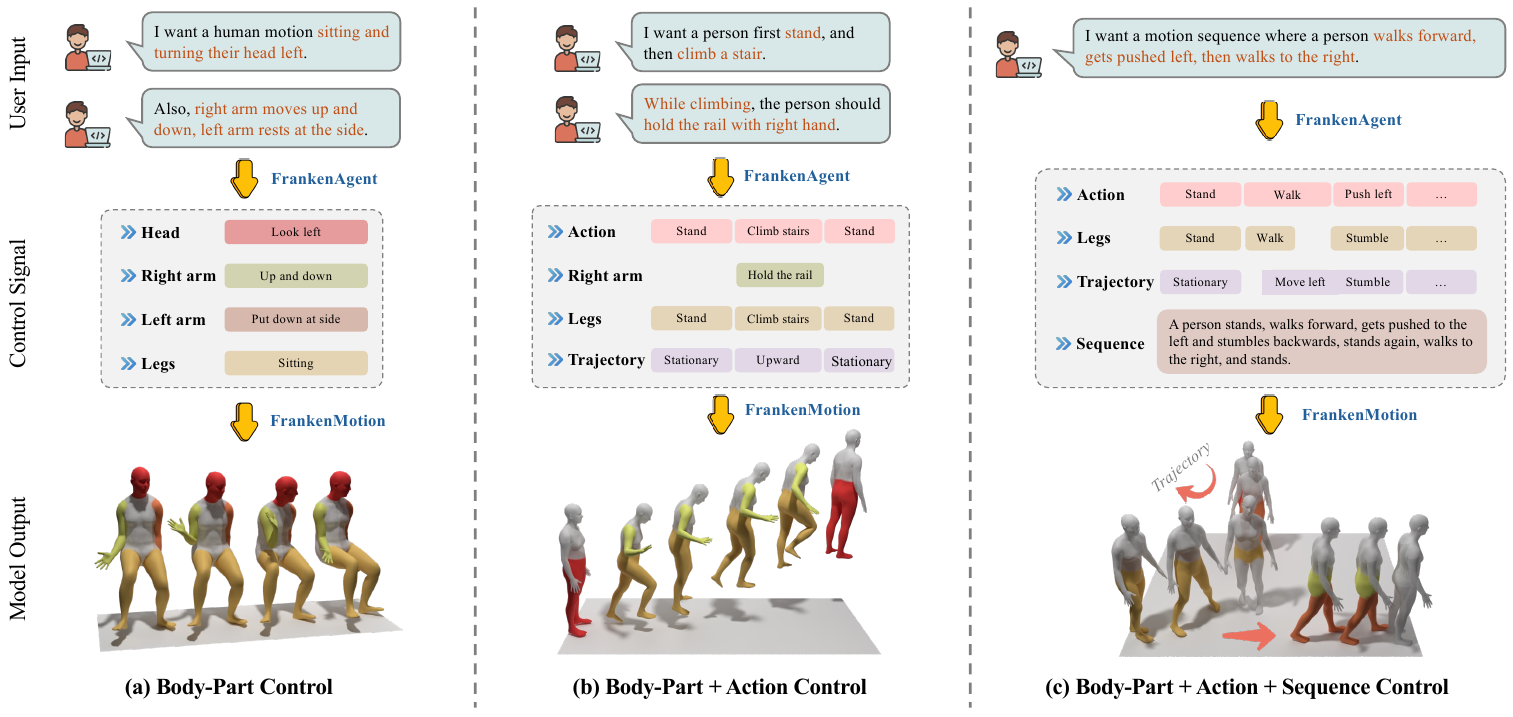}
    \vspace{-3mm}
    \captionof{figure}{Overview of our \textbf{\methodName{}} framework.
    \textbf{Left}: Body-Part Control, where users specify fine-grained movements of individual body parts; \textbf{Middle}: Body-Part + Action Control, enabling coordinated whole-body actions with part-specific constraints; \textbf{Right}: Body-Part + Action + Sequence Control, supporting complex multi-stage motion sequences involving interactions and transitions. In all cases, \textbf{\llmName{}} translates natural-language instructions into structured control signals for precise motion generation.}
    \label{fig:teaser_figure}
\end{center}
}]

\input{sections/0_abstract}

\input{sections/1_intro}

\input{sections/2_related}

\input{sections/3_dataset}
\input{sections/3_method}

\input{sections/4_experiment}

\input{sections/5_conclusion}

{
    \small
    \bibliographystyle{ieeenat_fullname}
    \bibliography{main}
}

\end{document}

%% file: math_commands.tex
\usepackage{amsmath,amsfonts,bm}

\def\eqref#1{equation~\ref{#1}}

\def\1{\bm{1}}

\DeclareMathAlphabet{\mathsfit}{\encodingdefault}{\sfdefault}{m}{sl}
\SetMathAlphabet{\mathsfit}{bold}{\encodingdefault}{\sfdefault}{bx}{n}

%% file: definitions.tex
\newcommand{\methodName}{FrankenMotion}
\newcommand{\dataName}{FrankenStein}
\newcommand{\llmName}{FrankenAgent}

\newcommand{\mat}[1]{\mathbf{#1}}
\newcommand{\set}[1]{\mathcal{#1}}
\newcommand{\vect}[1]{\mathbf{#1}}

\AtEndPreamble{
    \usepackage[capitalize]{cleveref}
    \crefname{section}{Sec.}{Secs.}
    \Crefname{section}{Section}{Sections}
    \Crefname{table}{Table}{Tables}
    \crefname{table}{Tab.}{Tabs.}
}

%% file: sections/0_abstract.tex
\begin{abstract}
Human motion generation from text prompts has made remarkable progress in recent years. However, existing methods primarily rely on either sequence-level or action-level descriptions due to the absence of fine-grained, part-level motion annotations. This limits their controllability over individual body parts.
In this work, we construct a high-quality motion dataset with atomic, temporally-aware part-level text annotations, leveraging the reasoning capabilities of large language models (LLMs). Unlike prior datasets that either provide synchronized part captions with fixed time segments or rely solely on global sequence labels, our dataset captures asynchronous and semantically distinct part movements at fine temporal resolution.
Based on this dataset, we introduce a diffusion-based part-aware motion generation framework, namely \methodName{}, where each body part is guided by its own temporally-structured textual prompt. 
This is, to our knowledge, the first work to provide atomic, temporally-aware part-level motion annotations and have a model that allows motion generation with both spatial (body part) and temporal (atomic action) control. 
Experiments demonstrate that \methodName{} outperforms all previous baseline models adapted and retrained for our setting, and our model can compose motions unseen during training. 
Our code and dataset will be publicly available upon publication.
\end{abstract}

%% file: sections/1_intro.tex
\section{Introduction}
Human motion generation is a fundamental task with broad applications in augmented reality (AR) and virtual reality (VR), gaming, entertainment, and embodied AI~\cite{zhang2024motion,zhu2023human}. 
In recent years, significant progress has been achieved in this field, largely driven by the growing availability of motion capture (mocap) datasets and their corresponding textual annotations.~\cite{Guo2022CVPR_humanml3d,plappert2016kit,mahmood2019amass, huang2024como}. These advancements have enabled motion generation models conditioned on various modalities, including text, scene layouts, and object interactions~\cite{motionlcm,li2024unimotion,xu2023interdiff,jiang2024scaling}. Concurrently, various extensions of the base task~\cite{braun2024physically,zhang2024force}, including motion editing, physics-aware generation, and style transfer, have been introduced. 

However, existing methods struggle to achieve both temporal and part-level control, primarily due to limitations in both dataset annotation and model design. For example, models trained on mocap datasets can generate realistic motions~\cite{tevet2023human, huang2024como, jiang2024motiongpt, guo2023momask}, yet they still lack fine-grained control over temporal dynamics and body parts. This is because most mocap datasets~\cite{amass} lack temporally and part-level aligned annotations. To address this, several approaches~\cite{lin2023motionx,zhang2024finemogen} have incorporated part-level labels to enhance model performance. Despite leveraging such information, these methods still fail to capture temporally coherent part-level features, which are crucial for fine-grained motion generation and controllable editing.

In this paper, we tackle motion generation from the perspective of composition. We argue that complex motions can be decomposed into simpler atomic motion elements. The key to allowing fine-grained control and complex motion generation lies in designing models that can learn these fundamental motion elements and their compositional relationships. We consider body parts as the basic elements in motion composition and design a model that maps prompts of body parts into complete temporal segments, the atoms of motion. We also input the high level action descriptions so that the model understands how different body parts compose into semantically meaningful complex motions.

Training such a model requires detailed per frame body part annotation, which is prohibitive to obtain. There exist datasets with textual annotations yet they lack structured part annotations. 
HumanML3D~\cite{Guo2022CVPR_humanml3d} and KITML~\cite{plappert2016kit} feature high level summary of a full motion sequence. Babel~\cite{BABEL:CVPR:2021} contains action level annotations such as `walk', `stand', and `knock'. Some actions, like `raising arms', do describe what one part is doing yet the annotations are not structured, and the part movement of most actions, such as `knock', remains unannotated. However, if a person sits down, the knees are probably bending; if a person is tying their shoes, their spine is bending, and the arms are tying the shoelaces. 
In other words, we can infer what the body parts are doing from high-level descriptions. Remarkably, we find that LLMs are powerful in inferring such relationships. Thus, we instantiate \llmName{}, an LLM agent that consumes existing datasets and outputs coherent per-frame body part annotations together with high level annotations. Using this, we automatically annotate the largest part-level human motion dataset to this date, which we call \dataName{} dataset.

Using \dataName{}, we train \methodName{}, a method that can be controlled at the sequence, action and part level as illustrated in~\cref{fig:teaser_figure}. By constitutionality, we can create novel movements not seen during training, such as a person sitting while raising the left arm. Experiments show that our \llmName{} is highly reliable, with 93.08\% annotations considered correct by human experts.
Results on our annotated \dataName{} also show that our model consistently outperforms state-of-the-art methods on part based motion generation in terms of both semantic correctness and realism. 
In summary, our main contributions are:
\begin{itemize}
    \item We present \methodName{}, a text-to-motion model that learns to compose complex motions through hierarchical conditioning on part-, action-, and sequence-level text, enabling fine-grained control over body parts and timing.
    \item We introduce \dataName{}, a new dataset with structured and temporally aligned body part annotations, leveraging existing datasets and LLM agent namely \llmName{}. 
    \item With hierarchical conditioning, our model allows flexible controlling at part, action or sequence level and generating novel motion compositions. 
\end{itemize}

%% file: sections/2_related.tex
\section{Related Work}

\noindent\textbf{Motion generation with control.}
Controllable motion generation has become increasingly important as it enables models to synthesize realistic motions aligned with user intent and environmental context. 
Among various modalities, text-conditioned motion generation~\cite{petrovich22temos,guo2023momask,Guo2022CVPR_humanml3d,shafir2023human,tevet2023human,motionlcm,zhang2022motiondiffuse,meng2024rethinking,Cho_2025_ICCV,Zhang_2025_CVPR,Zhao:DartControl:2025} has gained popularity for its flexibility and expressiveness. 
Other forms of control include inter-person conditioning~\cite{ruiz2024in2in,Wang_2025_CVPR,Ma_2025_ICCV,ruiz2025mixermdm} for modeling human–human interactions, key-pose or trajectory guidance~\cite{zhang2023motiongpt_aaai,bae2025less} for structural control, and audio-driven synthesis~\cite{genmo2025,zhang2025semtalk,li2025music} for cross-modal alignment. 
Human–object interaction modeling~\cite{zhang2022couch,Zeng_2025_CVPR,xu2024interdreamer,wu2025human,zhang2024hoi,xu2025intermimic,xu2025interact,zeng2025chainhoi,li2023object} further introduces affordance reasoning and contact-based control, while scene-aware methods~\cite{wang2025hsi,jiang2024scaling,wang2024move,pan2025tokenhsi} incorporate spatial constraints to ensure physical plausibility. 
Despite these advances, achieving fine-grained spatial and temporal control in motion generation remains a challenging open problem.

\noindent\textbf{Fine-grained Spatio-Temporal Motion Generation.}  
Achieving precise control in text-to-motion generation requires modeling both temporal and spatial structures.  
Existing approaches largely emphasize temporal control, regulating the duration and sequencing of motion segments at the frame level. Methods such as TEACH~\cite{TEACH:3DV:2022} compose motion segments using text and sequence-level conditioning, PriorMDM~\cite{shafir2023human} refines transitions between short clips via a two-stage inference process, and FlowMDM~\cite{barquero2024seamless} enhances temporal coherence by modeling smooth local transitions. DART~\cite{Zhao:DartControl:2025} employs a latent diffusion model for real-time, autoregressive motion generation, while UniMotion~\cite{li2024unimotion} introduces hierarchical control through both frame- and sequence-level conditioning. Beyond temporal modeling, spatial control has also gained attention. FineMoGen~\cite{zhang2024finemogen} supports diffusion-based motion generation and editing for fine-grained, per-body-part control, but its stage-based annotations enforce synchronized temporal intervals across parts, limiting flexibility. Besides, STMC~\cite{petrovich2024stmc} achieves spatial composition by assembling body-part motions from pretrained diffusion models, yet it remains a post-hoc method lacking end-to-end spatial–temporal reasoning.  While these methods produce impressive results, none provide unified, fine-grained multi-level control over atomic body parts, atomic actions, and sequence-level semantics along the temporal axis.

\begin{figure*}[t]
    \centering
    \includegraphics[width=1\linewidth]{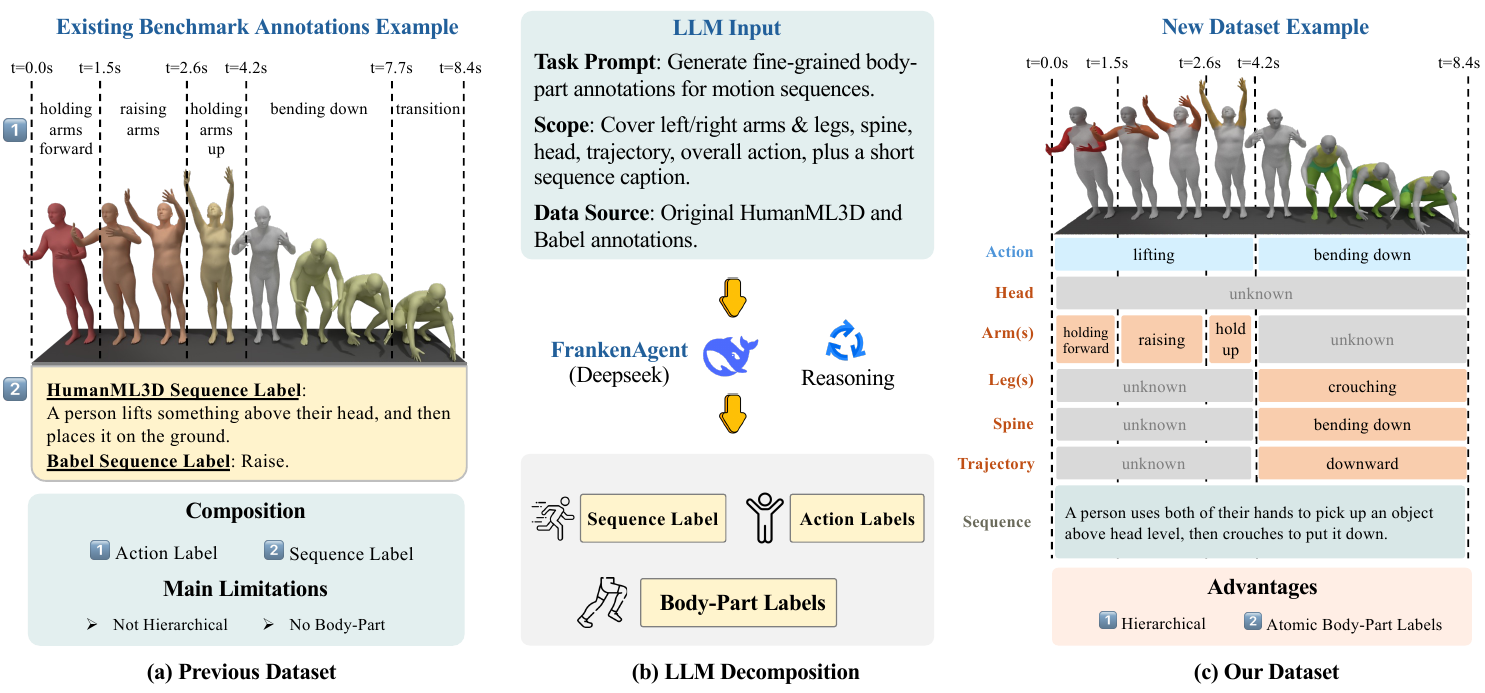}
    \caption{
    \textbf{LLM-assisted fine-grained motion annotation pipeline, compared with existing dataset.} Given motions with high level action descriptions, we instruct LLM to decompose the actions into part level descriptions and align with corresponding temporal windows. This gives the most important body part text and corresponding motions needed to learn essential motion elements. 
    }
    \label{fig:llm_annotation}
\end{figure*}

\noindent\textbf{Motion understanding and annotation.} Contrary to motion generation, motion understanding with natural language is also an important task for human behavior analysis and motion data annotation. Earlier work PoseScript~\cite{delmas2022posescript} generates text descriptions from pose parameters based on predefined rules. Follow up work MotionScript~\cite{yazdian2023motionscript} extends PoseScript to generate texts for motions programmatically. Despite detailed descriptions, the generated texts are complex and less similar to natural human language. Recent line of works focus more on using LLMs to understand human poses \cite{lin2025chathuman, feng2024chatpose} or motions~\cite{motionllm, jiang2024motiongpt, zhou2023avatargpt, mo2024mochat}. A common practice is to fine tune the pre-trained language models to align text and motion in the latent space. These methods differ in the formulation of motion understanding as language translation~\cite{jiang2024motiongpt} or interactive question answering~\cite{motionllm, mo2024mochat}. Some works can also handle raw videos~\cite{xu2024motionbank, Fang2025HuMoCon} in addition to 3D motions. In general, rule based annotation~\cite{delmas2022posescript, yazdian2023motionscript} lacks human readability, while fine-tuned models~\cite{jiang2024motiongpt} lose the generalization ability of original language models. Furthermore, these annotations focus on high-level action descriptions, lacking detailed body part decomposition to learn the essential motion elements. Our proposed annotation pipeline uses existing data and LLMs to generate atomic body-part level text prompts with temporal structure.

%% file: sections/3_dataset.tex
\section{\dataName{} Dataset Construction}
The quality and granularity of motion annotations directly constrain the upper bound of learning in motion understanding tasks. Existing datasets typically provide only coarse-grained labels, while lacking fine-grained temporal decomposition and body-part-specific annotations. For example, as shown in \cref{fig:llm_annotation}a, existing datasets only include sequence and action labels. To address this limitation, we propose a new annotation paradigm that explicitly operates at three levels of granularity: (1) \textbf{sequence level}: a global description of the full motion sequence, consistent with existing annotation formats; (2) \textbf{action level}: temporally localized coarse atomic actions; and (3) \textbf{body-part level}: fine-grained annotations for individual body parts (e.g., head, arms, legs, spine, trajectory) over time. This hierarchical annotation design enables a richer and more structured representation of human motion. As illustrated in \cref{fig:llm_annotation}b, the entire annotation process is automated, leveraging the reasoning capabilities of LLMs.

\subsection{Data Source}
Before presenting our automatic annotation pipeline, we first introduce several widely used motion–language datasets to which our approach can be applied: KIT-ML \citep{plappert2016kit}, BABEL \citep{BABEL:CVPR:2021}, AMASS~\cite{amass} and HumanML3D \citep{Guo2022CVPR_humanml3d}. KIT-ML is one of the earliest datasets linking motion capture data with natural language. It combines sequences from the CMU and KIT mocap datasets, providing motion–text pairs that enabled early research on motion–language alignment. However, it remains small in scale and lacks temporal segmentation or part-level annotations.
AMASS later unified motion capture data from 15 publicly available mocap datasets into a consistent parameterization of human motion, forming a large-scale foundation for subsequent motion–language datasets. Building on this resource, BABEL annotates a subset of AMASS with atomic action and coarse sequence-level labels, covering diverse activities but offering mainly categorical and short temporal annotations.
Similarly, HumanML3D extends AMASS with crowdsourced natural language descriptions, providing richer semantics but limited temporal resolution and no part-level detail.

Although these datasets have propelled progress at the intersection of motion understanding and natural language, they share two critical shortcomings: (1) \textbf{lack of hierarchical structure}: existing annotations provide only coarse action or sequence labels without temporally decomposing motions into sub-actions or body-part levels; and (2) \textbf{absence of body-part granularity}: the datasets do not specify which body parts are responsible for each motion, limiting fine-grained understanding.

\subsection{LLM-based Annotation Framework}
To address these limitations, we aim at building a motion dataset with structured text annotations that describe the motion at three \textit{different granularity levels} and are \textit{temporally aligned}. 
Specifically, given one motion sequence $\set{M}$ starts from $t=0$ and ends at $t=T$, we define one basic annotation element $a$ using a text label $L$ that describes the motion segment $m$ starts from $t_s$ and ends at $t_e$, formally: $a=(L, t_s, t_e)$. 
The goal is to obtain a structured collection of annotation elements $\set{A}=\{\set{A}_s, \set{A}_a, \set{A}_p\}$ that covers sequence level annotation $\set{A}_s$, atomic actions $\set{A}_a$, and body part annotations $\set{A}_p$. The sequence level annotation summarizes the full motion sequence in one text hence the annotation is simply one element: $\set{A}_s=\{(L_s, t_s=0, t_e=T)\}$. Atomic actions are a list of $N$ non-overlapping annotations, where each one summarizes a motion segment covering a temporal window: $\set{A}_a=\{(L^i, t_s^i, t_e^i)\}_{i=1}^N, \text{ with }t_s^1=0, t_e^N=T, \text{ and }t_s^{i}=t_e^{i-1}$. 
The part annotations, unique in our dataset, contain more fine-grained and structured annotations for $K$ body parts: $\set{A}_p=\{\set{A}_k\}_{k=1}^{K}$, here $\set{A}_k$, similar to atomic actions $\set{A}_a$, contains atomic motion annotations but with description specific for body part $k$. Formally: $\set{A}_k=\{(L^j_k, t_s^j, t_e^j)\}_{j=1}^M$. Note that annotating every single element in this collection $\set{A}$ is too expensive and unnecessary. Hence we allow the text labels $L$ to be \emph{unknown} for atomic actions $\set{A}_a$ and part annotations $\set{A}_p$, see example annotation in \cref{fig:llm_annotation}c. 

\begin{table}[t]
\centering
\resizebox{0.49\textwidth}{!}{
\begin{tabular}{l|ccc|c}
\toprule
\textbf{Attributes} & BABEL  & HumanML3D & KITML & \textbf{Ours} \\
\midrule
\multicolumn{5}{c}{\textit{Annotation Type}} \\ \midrule
Sequence Label      & \checkmark & \checkmark & \checkmark &  \checkmark \\
Atomic Action Label      & \checkmark & \xmark & \xmark &  \checkmark \\
Body-part Label & \xmark & \xmark & \xmark & \checkmark \\ \midrule
\multicolumn{5}{c}{\textit{Statistics}} \\ \midrule
Dataset Size & 43.5h & 28.6h & 11.2h & 39.1h \\
Vocabulary & 2,162 & 6,995 & 1,577 & 4,117 \\
Total Label  & 91.4k & 44.9k & 6.3k & 138.5k\\ \midrule 
Unseen Label & \multicolumn{3}{c|}{N/A}  & 28.8k \\ 
Part Label & \multicolumn{3}{c |}{N/A}   & 46.1k \\ 
\bottomrule
\end{tabular}}
\caption{Comparison of source motion--language datasets and our extended dataset. 
Our dataset builds upon \textbf{KIT-ML}~\cite{plappert2016kit}, 
\textbf{BABEL}~\cite{punnakkal2021babel}, and 
\textbf{HumanML3D}~\cite{Guo2022CVPR_humanml3d}, using LLM-based reasoning to produce multi-level, part-aware, and unseen annotations.}
\label{tab:dataset_comparison}
\end{table}

Leveraging the powerful body part reasoning capability of LLMs, we instantiate \llmName{} to construct body part annotations $\set{A}_p$ from sequence annotation $\hat{\set{A}}_s$ and atomic actions $\hat{\set{A}}_a$ provided by existing datasets. To ensure consistency with part annotations, we also further refine the existing sequence and part annotations: 
\begin{equation}
    \mathcal{A}_p, \mathcal{A}_a, \mathcal{A}_s = \text{\llmName{}}(\hat{\mathcal{A}}_a, \hat{\mathcal{A}}_s)
    \label{eq:franken-agent}
\end{equation}

We adopt Deepseek-R1 as our primary \llmName{} due to its strong reasoning ability and robust long-context understanding\footnote{We use Deepseek-R1-0528: \url{https://www.deepseek.com/}.}. We carefully design prompts to ensure high-quality decomposition. The LLMs are instructed to provide temporally aligned annotations with explicit body-part coverage, to decompose complex actions into interpretable segments, and to avoid vague expressions by outputting unknown when uncertain. BABEL serves as the primary reference, with KIT-ML and HumanML3D used for augmentation when available. To avoid hallucination, we explicitly instruct the agent to produce \textit{unknown} for the motion segments that it is unsure.

Detailed prompt templates can be found in Supp.
\subsection{Comparison with Existing Datasets}

\begin{figure*}[t]
    \centering
    \includegraphics[width=1\linewidth]{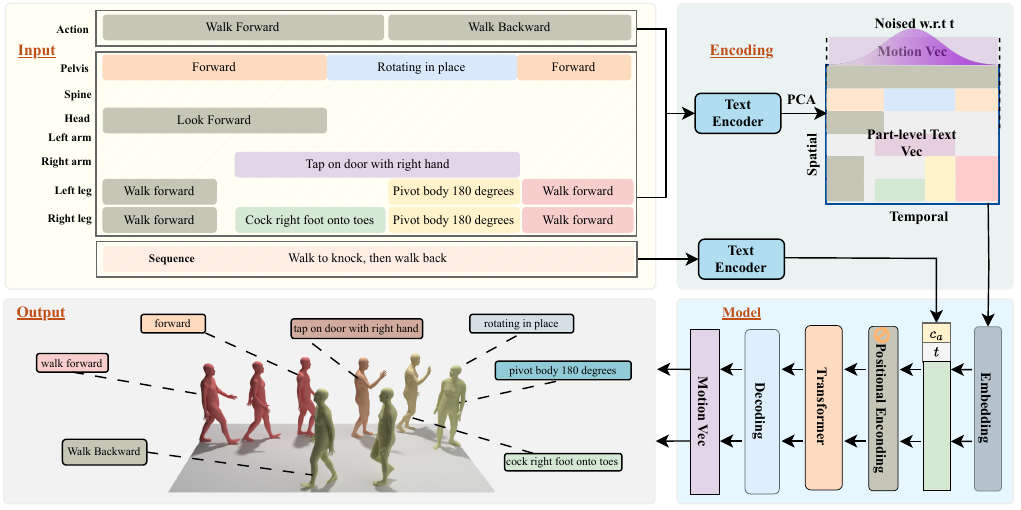}
    \vspace{-5mm}
    \caption{
    \textbf{Overview of our \methodName{}}.  
    Our model is a transformer-based diffusion model that can be input conditioned on a) sequence level prompt, b) action-level prompt and c) part-level prompt. After training with our paired data of motion and structured multi-granularity text annotations, it learns the essential motion elements and how to compose them into complex motions.
    }
    \label{fig:model}
\end{figure*}

 Based on our pipeline, we obtain a large-scale, high-quality dataset that substantially improves over existing datasets, see comparison in~\cref{tab:dataset_comparison}. Our dataset offers two major advantages: (1) \textbf{Hierarchical}: it provides multi-level annotations that span from the overall sequence to individual body parts; and (2) \textbf{Atomic labels}: each body part and atomic actions are annotated.

Statistically, our dataset spans 39 hours of motion data and includes three levels of labels with around 15.7k, 31.5k, and 46.1k annotations respectively, forming approximately 16k sequences and 265k atomic motion segments with an average duration of 4.8 seconds. We split the dataset into 80\% training, 10\% validation, and 10\% testing subsets. 
Beyond human-provided annotations, our system infers 28.8k new annotations through reasoning based on existing ones. \dataName{} offers precise spatial–temporal grounding at the atomic level, making it both challenging and valuable for advancing fine-grained motion understanding and generation. Further details are provided in the Supp. 

%% file: sections/3_method.tex
\section{\methodName{}: Part-Based Spatiotemporal Composition}
We aim to build a model that learns the spatial and temporal relationship between different parts and high level action semantics, allowing complex motion composition and fine-grained input control and editing. 
In the following, we start with problem setup (\cref{subsec:preliminary}) and then discuss our network details (\cref{subsec:spatiotemporal}) followed by robust training strategy~\cref{subsec:robust-train}). An overview can be found in \cref{fig:model}. 

\subsection{Preliminaries}\label{subsec:preliminary}

\noindent\textbf{Pose representation.}
We follow STMC~\cite{petrovich2024stmc} to represent a single frame pose $\vect{x} \in \mathbb{R}^d$ using SMPL~\cite{smpl} pose parameters, joint positions, velocities and angular velocities: 
\begin{equation}
\vect{x} = [r_{z}, \dot{r}_{x}, \dot{r}_{y}, \dot{\alpha}, \boldsymbol{\theta}, \mathbf{j}],
\label{eq:pose-representation}
\end{equation}
\noindent where $r_{z}$ is the $Z$ ($\uparrow$) coordinate of the pelvis, $\dot{r}_{x}$ and $\dot{r}_{y}$ are the linear velocities of the pelvis in the $x$–$y$ plane, and $\dot{\alpha}$ is the angular velocity around the vertical ($Z$) axis. 
$\boldsymbol{\theta}$ denotes the SMPL pose parameters (encoded using the 6D representation~\cite{zhou2019continuity}), and $\mathbf{j}$ are the 3D joint positions computed from the SMPL layer, forming a rotation-invariant representation by defining $\mathbf{j}$ in a local coordinate frame aligned with the body. 
To make $\boldsymbol{\theta}$ local to the body, the $Z$ rotation is removed from the SMPL global orientation. 

\noindent\textbf{Multi-granularity control.}
We adopt a transformer-based diffusion model as the framework for our text conditioned motion generation. To generate one motion sequence of $T$ frames, our model allows text control at three different levels of granularity: 1). $\mat{L}_s=\{L_s\}$: a single text description for the full sequence. 2). $\mathbf{L}_a=\{L_a^j | j\in\{1...W\} \}$: action labels of $W$ different non-overlapping temporal windows for the full sequence. %
3). Body part prompts $\mat{L}_p = \{L_k^i|i\in \{1,...T\}, k\in \{1,...,K\}\}$: text prompt for body part $k$ (among $K$ predefined parts) at each frame $i$. This is the most fine-grained input condition and allows part-based motion composition and editing. At inference time, users can conveniently provide only a sequence-level description, specify sparse part-level prompts, or edit existing control signals to guide the generated motion.

For the output, we adopt the pose representation defined in \cref{eq:pose-representation} aim at generating a sequence of realistic human motion: $\mathbf{x}^{[1...T]}$.
We adopt the sample prediction mode in diffusion models. Specifically, 
let $\mathbf{x}_0^{[1...T]}$ be the clean motion sequence and $\mathbf{x}_\sigma^{[1...T]}$ be the noisy motion at diffusion step $\sigma$, our model $f_\theta$ predicts the clean motion from three levels of text prompts $\mathbf{L}_s, \mathbf{L}_a, \mathbf{L}_p$ and diffusion timestep $\sigma$:
\begin{equation}
    \hat{\mathbf{x}}_0^{[1...T]}=f_\theta(\mathbf{x}_\sigma^{[1...T]}, \sigma, \mathbf{L}_s, \mathbf{L}_a, \mathbf{L}_p)
    \label{eq:model-prediction}
\end{equation}
To effectively learn the complex structured information of sequence, action and part level input, we design a network that extracts the features from different granularity levels into a joint latent space which we discuss next. 

\subsection{Spatio-temporal Embedding} \label{subsec:spatiotemporal}

To effectively learn the atomic text-to-motion mapping and composition, we use a joint embedding for sequence, action, part-level text and motion, see Fig.~\ref{fig:model}. 

\noindent\textbf{Action-part-motion embedding:} we use CLIP~\cite{radford2021learning} to extract text features for all input prompts.
For action and part labels, we apply PCA to reduce the embedding dimension to $D=50$, yielding
$\mathbf{F}_a \in \mathbb{R}^{W \times D}$ and $\mathbf{F}_p \in \mathbb{R}^{T \times (K \times D)}$.
Each action embedding in $\mathbf{F}_a$ corresponds to one of the $W$ temporal windows.To align these with the per-frame features, we expand each window embedding to its corresponding frame range, producing 
$\mathbf{F}_a \in \mathbb{R}^{T \times D}$. 
Thus, every frame is associated with both detailed part-level and high-level action text features, 
where $D_{m+t}$ denotes the dimension of the fused motion–text feature space.
We then concatenate both matrices, leading to a matrix of text features $\mathbf{F}_{a+p}\in \mathbb{R}^{N\times (K+1)D}$. 
The combined text feature $\mathbf{F}_{a+p}$ is then further concatenated with the noisy motion input $\mathbf{x}_\sigma^{[1...T]}$ to form the full conditioning input for the model. After passing through an MLP, this fusion produces motion–text embeddings $\mathbf{F}_{a+p+m} \in \mathbb{R}^{T \times D_{m+t}}$.  

\noindent\textbf{Sequence level context: }To incorporate sequence-level context, we encode the sequence text $\mathcal{C}_s$ using the CLIP text encoder followed by an MLP, obtaining a global feature vector $\mathbf{F}_s \in \mathbb{R}^{D_{m+t}}$. 
We append this global feature as an additional token to the fused sequence representation. 
The diffusion timestep embedding, after an MLP projection, is also added as a separate token. 
Together, these yield a final input representation of size $\mathbb{R}^{(T+2) \times D_{m+t}}$, which jointly encodes motion, part-level, action-level, sequence-level, and diffusion timestep information.

\subsection{Robust \methodName{} Training}\label{subsec:robust-train}

\noindent\textbf{Masking strategy.}  
We set zero to our text feature when text label is \textit{unknown}. This creates sparse input text condition as our body part annotations are sparse. To improve robustness, we introduce random masking with stochastic masking probability for each labelled text condition. We adopt Beta distribution to randomly decide the zero out probability $p$ of a body part text label $L_k^i$: $p\sim \mathrm{Beta}(5r, 5(1-r))$, where $r$ is the desired masking rate. At each training step, we sample different $p$ for body part labels $L_k^i$ that are not \emph{unknown}. This stochastic masking enhances robustness to incomplete conditioning and improves generalization under sparse supervision~\cite{Liu2019BetaDropout}.

\noindent\textbf{Training loss.} We train our diffusion model $f_\theta$ parametrized by $\theta$ using the standard DDPM objective~\cite{Ho2020DDPM}:
\begin{equation}
    \mathcal{L} = \mathbb{E}_{\substack{\mathbf{x}_0^{[1...T]},\, \sigma, \mathbf{\epsilon}}} \Bigl[\| f_\theta(\mathbf{x}_\sigma^{[1...T]}, \sigma, \mathbf{c}) - \mathbf{x}_0 \|_2^2\Bigr]
    \label{eq:ddpm-loss}
\end{equation}
where $\boldsymbol{\epsilon}\sim\mathcal{N}(0,\mathbf{I})$ is the diffusion noise and $\mathbf{c}=(\mathbf{L}_s, \mathbf{L}_a, \mathbf{L}_p)$ is our hierarchical text condition.

\noindent\textbf{Implementation details.}  
Following~\cite{petrovich23tmr}, we employ a cosine noise schedule with 100 diffusion steps, as introduced by~\cite{chen2023importance}. 
We use the AdamW optimizer~\cite{loshchilov2017decoupled} with a learning rate of $2\times10^{-4}$ and a batch size of 32. 
For text encoding, we adopt the frozen text encoder from CLIP (ViT-B/32)~\cite{radford2021learning}. 
Our model trains for approximately 47.5 hours on a single NVIDIA H100 GPU and each evaluation model trains for around 16 hours on a single NVIDIA A100 GPU.

%% file: sections/4_experiment.tex
\begin{table*}[t]
\centering
\scriptsize
\setlength{\tabcolsep}{3.5pt} %
\renewcommand{\arraystretch}{0.9} %

\begin{adjustbox}{center, max width=0.84\textwidth}
\begin{tabular}{l *{3}{c} *{3}{c} *{3}{c} *{2}{c} *{2}{c}}
\toprule
\multirow{2}{*}{Method} &
\multicolumn{3}{c}{\shortstack{Avg-part\\semantic correctness}} &
\multicolumn{3}{c}{\shortstack{Per-action\\semantic correctness}} &
\multicolumn{3}{c}{\shortstack{Per-seq\\semantic correctness}} &
\multicolumn{2}{c}{\shortstack{Per-action\\realism}} &
\multicolumn{2}{c}{\shortstack{Per-seq\\realism}} \\
\cmidrule(lr){2-4} \cmidrule(lr){5-7} \cmidrule(lr){8-10} \cmidrule(lr){11-12} \cmidrule(lr){13-14}
& R@1 $\uparrow$ & R@3 $\uparrow$ & M2T $\uparrow$
& R@1 $\uparrow$ & R@3 $\uparrow$ & M2T $\uparrow$
& R@1 $\uparrow$ & R@3 $\uparrow$ & M2T $\uparrow$
& FID $\downarrow$ & Div. $\rightarrow$
& FID $\downarrow$ & Div. $\rightarrow$ \\
\midrule

GT
& $52.04^{\pm 0.16}$ & $64.88^{\pm 0.18}$ & $0.71^{\pm 0.00}$
& $54.83^{\pm 0.12}$ & $72.42^{\pm 0.11}$ & $0.77^{\pm 0.00}$
& $72.66^{\pm 0.19}$ & $91.47^{\pm 0.17}$ & $0.78^{\pm 0.00}$
& $0.00^{\pm 0.00}$ & $53.56^{\pm 0.05}$
& $0.00^{\pm 0.00}$ & $48.81^{\pm 0.04}$ \\ \midrule

STMC
& $40.67^{\pm0.29}$ & $51.38^{\pm 0.30}$ & $0.66^{\pm 0.00}$
& $40.96^{\pm 0.61}$ & $56.32^{\pm 0.66}$ & $0.70^{\pm 0.00}$
& $43.58^{\pm 0.35}$ & $62.32^{\pm 0.45}$ & $0.67^{\pm 0.00}$
& $0.10^{\pm 0.00}$ & $51.79^{\pm 0.10}$
& $0.20^{\pm 0.00}$ & $46.82^{\pm 0.17}$ \\

DartControl
& $38.67^{\pm0.70}$ & $50.23^{\pm0.46}$ & $0.65^{\pm0.00}$
& $39.77^{\pm0.77}$ & $57.55^{\pm0.46}$ & $0.69^{\pm0.00}$
& $54.28^{\pm0.60}$ & $76.95^{\pm0.68}$ & $0.70^{\pm0.00}$
& $0.14^{\pm0.00}$ & $52.66^{\pm0.21}$
& $0.28^{\pm0.00}$ & $46.58^{\pm0.03}$ \\

UniMotion
& $45.72^{\pm0.24}$ & $57.36^{\pm 0.20}$ & $\textbf{0.69}^{\pm 0.00}$
& $47.58^{\pm 0.12}$ & $65.62^{\pm 0.33}$ & $\textbf{0.75}^{\pm 0.00}$
& $62.66^{\pm 0.30}$ & $82.08^{\pm 0.35}$ & $0.74^{\pm 0.00}$
& $0.05^{\pm 0.00}$ & $53.12^{\pm 0.23}$
& $0.08^{\pm 0.00}$ & $48.36^{\pm 0.03}$ \\

\midrule

\methodName{}
& $\textbf{47.21}^{\pm 0.19}$ & $\textbf{58.97}^{\pm 0.18}$ & $\textbf{0.69}^{\pm 0.00}$
& $\textbf{48.10}^{\pm 0.13}$ & $\textbf{65.79}^{\pm 0.14}$ & $\textbf{0.75}^{\pm 0.00}$
& $\textbf{65.27}^{\pm 0.22}$ & $\textbf{85.62}^{\pm 0.18}$ & $\textbf{0.76}^{\pm 0.00}$
& $\textbf{0.04}^{\pm 0.00}$ & $\textbf{53.82}^{\pm 0.10}$
& $\textbf{0.06}^{\pm 0.00}$ & $\textbf{48.60}^{\pm 0.05}$ \\

\bottomrule
\end{tabular}
\end{adjustbox}

\vspace{-2mm}
\caption{
\textbf{Evaluating text to motion generation.} We report the semantic correctness and realism of parts (averaged), action and sequence level motion, with 95\% confidence interval ($\pm$) after 20 repeated evaluations. 
Across all settings, our \methodName{} achieves the best performance, outperforming all prior baselines in both correctness and realism.
}
\label{tab:baselines}
\end{table*}

\begin{figure*}[ht]
    \centering
    \includegraphics[width=\textwidth]{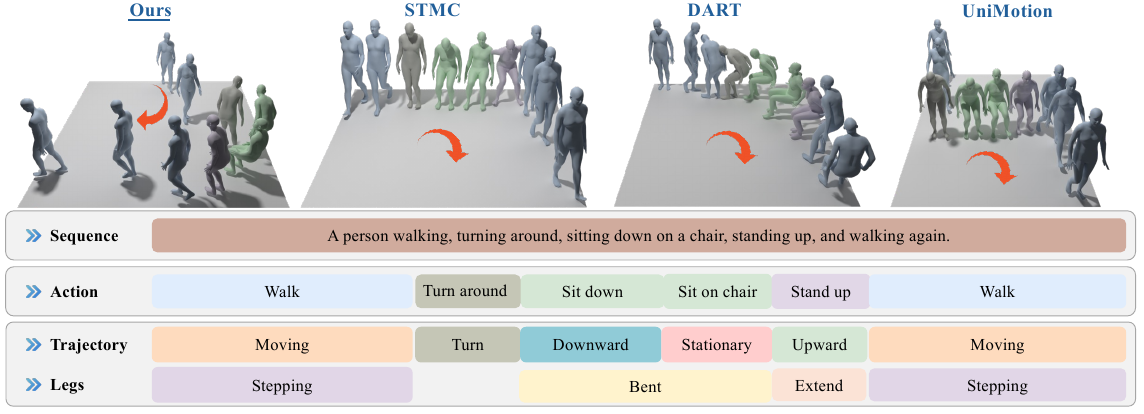}
    \caption{
    \textbf{Qualitative comparison with baselines}. Prior methods cannot compose parts into realistic motion (STMC~\cite{petrovich2024stmc}), generates repetitive motions (DART~\cite{Zhao:DartControl:2025}), or do not follow the intricate details like ``turn around'' (UniMotion~\cite{li2024unimotion}). Our method faithfully composes the complex parts into one realistic motion while also following precisely the detailed body part prompts and high level semantics. 
    }
    \label{fig:baseline}
\end{figure*}

\begin{table*}[t]
\centering
\scriptsize
\setlength{\tabcolsep}{3.4pt}
\renewcommand{\arraystretch}{0.95}

\begin{adjustbox}{center, max width=0.91\textwidth}
\begin{tabular}{l ccc ccc ccc ccc cc}
\toprule
\multirow{2}{*}{Method} &
\multicolumn{3}{c}{Inputs} &
\multicolumn{3}{c}{\shortstack{Avg-part\\semantic correctness}} &
\multicolumn{3}{c}{\shortstack{Per-action\\semantic correctness}} &
\multicolumn{3}{c}{\shortstack{Per-seq\\semantic correctness}} &
\multicolumn{2}{c}{Realism} \\
\cmidrule(lr){2-4}\cmidrule(lr){5-7}\cmidrule(lr){8-10}\cmidrule(lr){11-13}\cmidrule(lr){14-15}
& Part & Atomic & Seq. &
R@3 $\uparrow$ & M2T $\uparrow$ & M2M $\uparrow$ &
R@3 $\uparrow$ & M2T $\uparrow$ & M2M $\uparrow$ &
R@3 $\uparrow$ & M2T $\uparrow$ & M2M $\uparrow$ &
FID $\downarrow$ & Div. $\rightarrow$ \\
\midrule

\multirow{1}{*}{GT}
& \cmark & \cmark & \cmark &
$64.88^{\pm 0.18}$ & $0.71^{\pm 0.00}$ & $1.00^{\pm 0.00}$ &
$72.42^{\pm 0.11}$ & $0.77^{\pm 0.00}$ & $1.00^{\pm 0.00}$ &
$91.47^{\pm 0.17}$ & $0.78^{\pm 0.00}$ & $1.00^{\pm 0.00}$ &
$0.00^{\pm 0.00}$ & $46.84^{\pm 0.03}$ \\

\midrule
\multirow{3}{*}{Ours}
& \cmark & \xmark & \xmark &
$56.34^{\pm 0.42}$ & $0.69^{\pm 0.00}$ & $0.72^{\pm 0.00}$ &
$--$ & $--$ & $--$ &
$--$ & $--$ & $--$ &
$0.08^{\pm 0.00}$ & $46.09^{\pm 0.09}$ \\

& \cmark & \cmark & \xmark &
$57.74^{\pm 0.42}$ & $0.69^{\pm 0.00}$ & $0.73^{\pm 0.00}$ &
$65.39^{\pm 0.17}$ & $0.75^{\pm 0.00}$ & $0.75^{\pm 0.00}$ &
$--$ & $--$ & $--$ &
$0.07^{\pm 0.00}$ & $46.19^{\pm 0.07}$ \\

& \cmark & \cmark & \cmark &
$\textbf{58.97}^{\pm 0.18}$ & $0.69^{\pm 0.00}$ & $\textbf{0.75}^{\pm 0.00}$ &
$\textbf{65.79}^{\pm 0.14}$ & $0.75^{\pm 0.00}$ & $\textbf{0.76}^{\pm 0.00}$ &
$\textbf{85.62}^{\pm 0.18}$ & $\textbf{0.76}^{\pm 0.00}$ & $\textbf{0.75}^{\pm 0.00}$ &
$\textbf{0.05}^{\pm 0.00}$ & $\textbf{46.53}^{\pm 0.03}$ \\

\bottomrule
\end{tabular}
\end{adjustbox}

\caption{
\textbf{Importance of hierarchical input condition.} We train models that consume input of part only, or additionally with atomic action, or all part, action and sequence level (Seq.) texts. Even with part-level inputs alone, our model attains strong performance, achieving an M2T score close to the upper-bound GT reference. Incorporating action and sequence texts introduces high-level semantics for the desired motion, which further enhances both the correctness and realism of part-level motion generation. 
}
\label{tab:ablation}
\end{table*}

\section{Experiments}
In this section, we first evaluate our newly introduced motion dataset, and then compare our model with baselines for hierarchical text to motion generation. We further ablate the design choices of our model and showcase various applications due to the flexibility of our model design. Results validate the high quality of our data and demonstrate that our model outperforms all previous methods.

\subsection{Dataset Quality}\label{subsec:data-quality}
To evaluate the quality of our \llmName{}, we conduct a human evaluation on the annotated \dataName{} dataset. We randomly sampled 50 motion sequences and asked three human experts to assess whether the generated part, action, and sequence labels are consistent with the corresponding motion segments. Each expert assigns a binary correctness score for every label, and we compute the average as the annotation accuracy. The overall accuracy of our \llmName{} generated annotation is 93.08\%, demonstrating the high accuracy of our method. We also report inter-annotator agreement using Gwet’s AC1 coefficient ($AC_1$)~\cite{gwet2001handbook} to assess the reliability of human evaluation, where we obtain a score of $AC_1$ = 0.91 ($0.8 \leq AC_1 \leq 1.0$), indicating high reliability. Please refer to the Supp for more details.

\subsection{Fine-grained Motion Generation}\label{subsec:baseline}
Our \methodName{} allows motion generation from hierarchical texts of sequence, atomic action, and body-part labels. To the best of our knowledge, none of the prior methods are able to accomplish this complex task. To be able to compare, we adapt prior methods to include part control and train on our proposed \dataName{} dataset. We outline the baseline setup below and provide the details in Supp.  

\noindent\textbf{Baselines.} We adapt UniMotion~\cite{li2024unimotion}, STMC~\cite{petrovich2024stmc}, and DART~\cite{Zhao:DartControl:2025} to our setup as the baselines. For fair comparison, all methods are modified to use the same pose state representation (\cref{eq:pose-representation}) defined in STMC~\cite{petrovich2024stmc} that includes SMPL~\cite{smpl} pose parameters, velocity, and joint positions. \textbf{1). STMC} is a post-hoc, test-time composition method that stitches body-part motions predicted by MDM~\cite{tevet2023human} at each frame. We retrain the base MDM on all possible motion-text pairs from our \dataName{} dataset to ensure that MDM understands our text labels. At inference time, STMC receives part labels as conditioning and stitches the part motions generated by the retrained MDM. For the frames where no part labels are given, we use atomic action labels or sequence level text.
\textbf{2). UniMotion} is a hierarchical model that supports sequence-level and frame-level (atomic action) text conditioning with temporal alignment but lacks body part control. We merge all part and action labels into one long text for each frame as the frame-level text input for UniMotion. 
\textbf{3). DART} is an autoregressive model that predicts future motion frames conditioned on motion history and a high-level text prompt. We merge our three levels of text into one long text for each frame and train DART with these pairs of text-motion data.

\noindent\textbf{Evaluation metrics.} Following~\cite{li2024unimotion}, we evaluate generated motions along two axes: \textbf{semantic correctness}, consisting of R-Precision~\cite{Guo2022CVPR_humanml3d} and M2T~\cite{petrovich2024stmc}, and \textbf{realism}, consisting of Fréchet Inception Distance (FID) and Diversity~\cite{Guo2022CVPR_humanml3d}. All these metrics require a pretrained text to motion generation model to measure the text-motion alignment. To assess the performance in all three levels of control, we train separate evaluation models for each input modality following TMR~\cite{petrovich23tmr}: seven for each body part, one for actions, and one for full sequences. Each model is trained with paired data of text and corresponding full body motion. For semantic correctness metrics, we follow~\cite{petrovich23tmr} and use MPNet~\cite{song2020mpnet} embeddings to remove false negative pairs due to paraphrased text labels (e.g., “hold arm lateral” vs. “hold arm horizontally”). Due to space constraints, we report semantic alignment scores for sequence-, action-, and averaged part-level evaluations, while realism metrics are reported for all action- and sequence-level crops. Additional numbers for part-level realism can be found in Supp.
For all metrics, we repeat the evaluation 20 times and report the 95\% confidence interval, indicated by the variance score.

\noindent\textbf{Results.} We compare our method against baselines quantitatively in \cref{tab:baselines} and qualitatively in \cref{fig:baseline}. STMC~\cite{petrovich2024stmc} follows part instructions thanks to the well trained MDM model, but it struggles to compose different parts into a realistic motion. This often leads to unsmooth transitions between atomic actions or fine-grained control, such as ``turning around'', being ignored, as can be seen in \cref{fig:baseline} column 2. UniMotion~\cite{li2024unimotion} generates more realistic motion due to its frame-level control, as can be seen from the FID score in \cref{tab:baselines}. However, it lacks an explicit structure of the body part features, leading to less precise text control, and the tiny detailed motions are ignored, as shown in the R-1 score (\cref{tab:ablation}) and \cref{fig:baseline}, where it does not turn around. DART primarily follows the sequence level text; however, it cannot control motion precisely at each frame. Due to its auto-regressive design, error can accumulate, and it often produces repeated motion segments representing only partial input prompts (see repeated sitting and standing in \cref{fig:baseline}).

In contrast, our model has structured body part conditioning, together with hierarchical control of atomic action and sequence level text. It generates fine grained motions that are controlled precisely by body parts while also maintaining coherence with the high level semantics introduced by atomic action and sequence level text. Notice in \cref{fig:baseline} column one how precisely our model follows the text prompt of body parts and actions at every frame. In \cref{tab:baselines}, our method clearly outperforms all baselines in both motion quality and consistency with the input text.

\subsection{Ablation Study}
We propose a model that consumes text at three levels of granularity, as we find that these different texts are important for generating coherent motions. 
To evaluate this, we train separate models that take only part text, part and atomic action, and all text as input conditions, and we evaluate the motion generation performance in \cref{tab:ablation}. We also show the scores of our evaluation model, which serve as an upper bound. Notably, the model conditioned only on part texts already achieves competitive scores, demonstrating the strong capability of our model for understanding fine-grained part texts. With additional high level semantics of atomic action and sequence level text, our model generates motions that more precisely follow the part labels. Furthermore, such motions are more meaningful due to the guidance of high level text semantics, illustrating the benefits of our hierarchical design. We show some qualitative examples in Supp.

\subsection{Flexible Input Control Applications}
Thanks to its modular design and the sparse structure of our dataset, our model supports flexible conditioning during inference. 
Users can control the motion at different granularities—such as a dominant body part, an action-level phrase, or a single sequence-level description—allowing adaptive control depending on the available text or user preference. %
We demonstrate the flexibility of our input in \cref{fig:teaser_figure} and \cref{fig:baseline}. 
Additional qualitative examples and more application cases are provided in the supplementary video.

%% file: sections/5_conclusion.tex
\section{Conclusion and Limitation}
We introduced multi-level spatiotemporal motion conditioning, enabling controllable generation at the sequence, atomic action, and atomic part levels.
To support this task, we built \dataName{}, a fine-grained, temporally aligned dataset with atomic part-level annotations derived via LLM reasoning.
Based on it, \methodName{} learns to compose motion from atomic elements, achieving fine-grained and flexible spatial–temporal control.
Experiments show that it outperforms adapted baselines and establishes a strong foundation for compositional motion generation.
\noindent\textbf{Limitation}
While \methodName{} enables fine-grained spatiotemporal text-to-motion generation, it cannot yet generate minute-long motion sequences within a single pass. Extending its ability to model long-term temporal structure will be an important direction for future work.
\clearpage
{\small
\noindent\textbf{Acknowledgments:} Special thanks RVH and AVG members for the help and discussion. Prof. Gerard Pons-Moll and Prof. Andreas Geiger are members of the Machine Learning Cluster of Excellence, EXC number 2064/1 - Project number 390727645. Gerard Pons-moll is endowed by the Carl Zeiss Foundation. Andreas Geiger was supported by the ERC Starting Grant LEGO-3D (850533). 
}